\documentclass{article}

\usepackage[final]{neurips_2022}

\usepackage{graphicx}
\usepackage{subfig}
\usepackage{algorithmic}
\usepackage{algorithm}
\usepackage{wrapfig}
\usepackage{lipsum}

\setcitestyle{numbers,square}

\usepackage{amsmath}
\usepackage{amssymb}

\renewcommand{\texttt}{}

\newcommand{\rnn}{\mathit{rnn}}

\newcommand{\rnnembed}{\mathit{E}^{\rnn}}

\def\E{I\!\!E}

\usepackage[utf8]{inputenc} 
\usepackage[T1]{fontenc}    
\usepackage{hyperref}       
\usepackage{url}            
\usepackage{booktabs}       
\usepackage{amsfonts}       
\usepackage{nicefrac}       
\usepackage{microtype}      
\usepackage{xcolor}         

\title{Explain My Surprise: Learning Efficient Long-Term Memory by Predicting Uncertain Outcomes}

\author{%
  Artyom Sorokin \\
  AIRI, MIPT\\
  Moscow, Russia\\
  \texttt{asorokin@airi.net} \\
  \And
   Nazar Buzun \\
   AIRI \\
   Moscow, Russia\\
   \texttt{buzun@airi.net} \\
   \And
   Leonid Pugachev\\
   MIPT \\
   Dolgoprudny, Russia \\
   \texttt{puleon@mail.ru}
   \And
   Mikhail Burtsev\\
   AIRI, MIPT \\
   Moscow, Russia\\
  \texttt{burtsev@airi.net} 
}

\begin{document}

\maketitle

\begin{abstract}
In many sequential tasks, a model needs to remember relevant events from the distant past to make correct predictions. Unfortunately, a straightforward application of gradient based training requires intermediate computations to be stored for every element of a sequence. This requires to store prohibitively large intermediate data if a sequence consists of thousands or even millions elements, and as a result, makes learning of very long-term dependencies infeasible. However, the majority of sequence elements can usually be predicted by taking into account only temporally local information. On the other hand, predictions affected by long-term dependencies are sparse and characterized by high uncertainty given only local information. We propose \texttt{MemUP}, a new training method that allows to learn long-term dependencies without backpropagating gradients through the whole sequence at a time. This method can potentially be  applied to any recurrent architecture.  LSTM network trained with \texttt{MemUP} performs better or comparable to baselines while requiring to store less intermediate data.  
\end{abstract}

\section{Introduction}
\label{sec:intro}

Two dominating approaches for memory augmentation in deep learning include recurrent neural networks~\cite{werbos1990backpropagation, lstm_hochreiter1997, graves2014neural, weston2014memory, a3c_mnih, memory_mincraft_oh2016, dnc_graves2016, relational_rnn_santoro2018, r2d2_kapturowski2018}) and Transformers \cite{transformers_vaswani2017, burtsev2020memory-transformer,lei-etal-2020-mart,gupta2020gmat,wu2020memformer,fan2020feedback-transformer}. 
Historically, recurrent networks and Transformers have been primarily developed for applications in natural language processing, which is characterized by strong dependencies between closely located elements. 
On the other hand, in the reinforcement learning setting, the agent often has to remember only a few bits of information but for a very long time~\cite{amrl_beck_2020}. 

Evolution of neural architectures tackling the problem of learning long-term dependencies follows the path of increasing size and complexity ~\cite{memory_mincraft_oh2016, relational_rnn_santoro2018, merlin_wayne2018}.
For example, Transformers\cite{stabilizing_transformer_2019} and memory augmented neural networks~\cite{dnc_graves2016} perform much better than the classical \texttt{LSTM} in tasks with long-term temporal dependencies. Still, even gigantic language models like GPT-3 \cite{brown2020language} process sequences only up to 2048 elements due to quadratic complexity of self-attention. 
So, these architectures are much more demanding in terms of the memory capacity and computational power required for training.

To learn a long-term dependency between a distant observation in the past and agent's actions in the current state, both the recurrent networks trained with back propagation through time and Transformers need to store gradients for all intermediate steps.
Taking into account, that the minimum distance between useful information in the past and the moment of its utilization can be measured in thousands or millions of time steps, such solutions seem unrealistic given the current hardware capabilities. 
We propose a training method that allows an agent to find and store dependencies between temporarily distant events without the need to process intermediate steps.

Our contributions are the following:
\begin{itemize}
    \item We propose a new formulation for the problem of memory learning as maximisation of conditional mutual information between memory state and future outcomes.
    \item We introduce an original method  \texttt{MemUP} (\textbf{Mem}ory for \textbf{U}ncertainty \textbf{P}rediction) for efficient long-term memory training by minimizing uncertainty of predictions  for states where local information is not enough.
    \item We show that implementation of our method for recurrent neural networks is able to learn long-term dependencies better than baselines while propagating gradients only through a small fractions of a sequence in one optimization step. Notably, the applications of MemUP are successful in supervised learning tasks with different loss functions as well as in reinforcement learning tasks.
\end{itemize}

\section{Memory to explain surprise}\label{sec:idea}


The main idea of our approach is very simple. Find the most surprising events or elements of a sequence and train a memory to detect and store information that allows to explain out these events. Below we formulate this idea in the framework of information theory.

Lets consider a sequence ${\tau}$ with inputs $\{x_t\}^T_{t=0}$ and targets $\{y_t\}^T_{t=0}$.
A model is trained to predict targets given inputs.
An ideal memory could just store all previous inputs  $\{x_i\}^t_{i=0}$ at each step $t$ and predict $y_t$ from them, but it is too computationally expensive. In practice a \texttt{memory model} $g_{\theta}$   has to selectively store some aggregation of this sequence in a memory state $m_t = g_{\theta}(x_t,m_{t-1})$
. Then it should be decided what to write and what to remove from the memory.

\paragraph{Estimating past information importance.}
Consider an arbitrary time step $k$.  The past element $x_{t}$ has useful information about the target $y_{k}$, if $y_{k}$ depends on $x_{t}$ given current $x_k$, i.e. $p(y_{k} |x_k, x_{t}) \ne p(y_{k} |x_k). $ 
 In this case, we can say that there is a temporal dependency with the length of $k-t$ steps, that starts at step ${t}$ and ends at step $k$. 
 The strength of temporal dependency or importance of remembering $x_{t}$ can be measured as an amount of change in the distribution of $y_k$ given a knowledge of $x_{t}$ . Multiplication of the both parts of the inequality above by $p(x_{t}| x_k)$ and application of KL-divergence as a measure of the difference between distributions leads to the conditional mutual information as a measure of $x_{t}$ importance:
\begin{equation}
I(y_{k} ;  x_{t}| x_k) = \E_{x_k } \mathcal{KL}[p(y_{k}, x_{t}|x_k) \|  p(y_{k}|x_k)p(x_{t}|x_k)],
\end{equation}
where $x_k$ is taken from sequences stored in some dataset or generated by an agent over interaction with the environment.
\paragraph{Objective function for memory training.}
Then, the problem of learning what to store in the memory can be defined as maximization of the mutual information between memory states and prediction targets with respect to the parameters~$\theta$:
\begin{equation}
    \label{eq:full_objective}
    \max_{\theta} 
    \sum^{T}_{k=t} I(y_{k} ; m_t = g_{\theta}(x_t, m_{t-1}) | x_k).
\end{equation}
The 
sum in eq. \ref{eq:full_objective} allows us to process each memory update in a separate gradient step. As accounting for all future steps prevents loss of information that can be helpful in the distant future. 
Unfortunately, this still requires the entire remaining (future) sequence to be processed for every update and has the complexity $O(T)$ with respect to the sequence length. 

\paragraph{Reducing the cost of memory training.}
Fortunately, updating the memory for all future steps is not necessary for the majority of real-world tasks. 
As the distance between events increases, the number of dependencies also decreases. This phenomenon is sometimes called locality of reference \cite{big_bird_zaheer2020}.
In other words, information from the greater part of past inputs may be useless for $y_k$ prediction given $x_k$.
Accordingly, most of the elements in the sum (eq. \ref{eq:full_objective}) will be close to zero.
Therefore, depending on the structure of the problem, we could vary the number of elements in the sum, leaving only those for which long-term memory is critical for correct predictions. The main problem is to find such steps in the sequence.

Lets assume that the ideal memory $m^*_t$ that stores all possible information from the past is available.  Then the conditional mutual information at a step $t$:
\begin{equation}
\label{eq:memory_potential}
I(y_t, m^*_t| x_t) = H(y_t|x_t) - H(y_t| m^*_t, x_t), 
\end{equation}
shows the potential utility of memory for prediction at this step. When the number of elements in the sum in eq. \ref{eq:full_objective} is minimized to reduce computations, the steps with the highest value of mutual information from eq. \ref{eq:memory_potential} should be preserved to assure good quality of prediction.  Naturally, we cannot estimate $I(y_t, m^*_t| x_t)$ directly, because $m^*_t$ is not available to us. 

As seen from the eq.\ref{eq:memory_potential}, $m^*_t$ participates only in the second entropy, which indicates how well one can predict the future in the presence of ideal memory and ideal model. On the other hand, the local entropy $H(y_t|x_t)$ can be easily estimated. A value of local entropy would be sufficient for our task if we could use it to determine the ordering for values of the potential memory utility $I(y_t, m^*_t|x_t)$. Specifically, for every pair of elements $i$, $j$ if $H(y_i|x_i) > H(y_j|x_j)$ then $I(y_i, m^*_i|x_i) > I(y_j, m^*_j|x_j)$. This becomes true under one of two possible conditions: (1) $H(y_t|x_t)\gg H(y_t|m^{*}_t, x_t)$ or (2) $H(y_t|m^{*}_t, x_t) \sim \epsilon$. The first condition means that given a perfect memory and a model, one could make high-quality predictions for the task at hand. The second condition means that the real distribution of the target variable $y_t$ has a similar amount of noise at each step.

If our task satisfies either one of these two conditions, then valuable information from the past could make the biggest contribution for the elements with the highest values of average "surprise" $H (y_t | x_t)$. 

We should note that estimation of $I(y_j, m^*_j|x_j)$ by the local entropy makes MemUP vulnerable to the Noisy-TV problem\cite{rnd_burda2018}, when the conditions we have described are not satisfied. Sensitivity tests for the Noisy-TV problem are discussed in Section \ref{sec:noisy_tv_problem_main}. 

Thus, instead of optimizing the memory for each future step, we introduce a new objective function that allows us to train it only for steps where the potential gain from the past information is maximal:
\begin{equation}
    \label{eq:cost_efficient_objective}
    \max_{\theta} 
    \sum_{k \in \mathcal{U}_t} I(y_{k} ; m_t = g_{\theta}(x_t, m_{t-1}) | x_k)\, ,
\end{equation}
here $\mathcal{U}_t$ denotes the set of top-$K$ indices of steps from $t$ to $T$ with the highest estimated local entropy $H (y_k | x_k)$. The hyperparameter $K$ controls fraction of a sequence to be processed in one gradient step. On the other hand, maximizing the mutual information between the memory $m_t$ and an arbitrary distant event at step $k > t$ allows to learn long-term dependencies.

\paragraph{Optimization.} Directly optimizing mutual information can be a challenging task. In our case, we use variational lower bound on mutual information proposed by \cite{im_agakov2004}. Using the lower bound we can maximize mutual information by minimizing Cross-Entropy (CE) between the empirical and model distribution for all selected high entropy events:
\begin{equation}
    \label{eq:surrogate_objective}
    \min_{\theta, \phi} 
    \sum_{k \in \mathcal{U}_t} \E_{x_k,y_k}\left[ - \log q_{\phi}(y_k|m^{\theta}_t, x_k)\right]\, ,
\end{equation}
where $m^{\theta}_t$ is a shortcut for $m_t = g_{\theta}(x_t, m_{t-1})$, $q_{\phi}$ is a \texttt{predictor} network with parameters $\phi$, that estimates probability of $y_k$ given $x_k$ and $m_t$. For a detailed derivation of eq. \ref{eq:surrogate_objective} see Supplementary Materials Section A. 

Collection of elements with high uncertainty $\mathcal{U}_t$ requires a special \texttt{uncertainty detector} $d_\psi$ model. The main requirement for $d_\psi$ is to produce uncertainty estimates $s_t= d_\psi(x_t)$.  In the simplest case, the detector estimates ``surprise'' $-\log p(y_t| x_t)$ (prediction error), which can be seen as a single point estimate of uncertainty. Other models that directly estimate  the uncertainty or a whole distribution \cite{Tsymbalov2020DropoutSB, qr_dqn_dabney2018} can also be used. 


\section{MemUP for Recurrent Neural Architectures}\label{subsec:rnn_impl}

Recurrent neural net implementation of \texttt{MemUP} consists of (1) training of uncertainty detector model $d_\psi$, (2) selection of elements with highest uncertainty $\mathcal{U}_t$ and (3) training memory $g_\theta$ and predictor  $q_\phi$ models.  A pseudocode for  \texttt{ RNN MemUP } training is shown in Supplementary Materials E.

\textbf{Uncertainty detector training.} Information maximization reasoning does not limit the choice of solutions for uncertainty detector $d_\psi$. In our experiments on algorithmic tasks we use a recurrent neural classifier trained with Cross-Entropy Loss. In this case detector's "surprise" $-\log\, d_\psi(y_t| x_t)$ is used as an uncertainty estimate.  

In Reinforcement Learning experiments we use distributional algorithm \texttt{QRDQN} \cite{qr_dqn_dabney2018} trained to predict discounted future returns $R_t$ ($y_t = R_t$ in RL experiments). As \texttt{QRDQN} approximates a whole distribution of the future returns, estimating uncertainty becomes a simple task.

\textbf{Processing a sequence with the memory network.} \texttt{ MemUP} allows to learn long term-dependencies without propagating gradients through the whole sequence. To demonstrate that in our experiments we use Truncated BPTT to train memory network $g_\theta$.
In the majority of  supervised learning experiments we set rollout length $r$ between 10 and 60 steps. In all Reinforcement Learning experiments $r=1$, i.e. recurrent memory is trained without actually using backpropagation through time. 
In All Experiments $g_\theta$ is a simple \texttt{LSTM} network with additional input encoder.

\begin{figure*}
\begin{center}
\centerline{\includegraphics[width=0.84\textwidth]{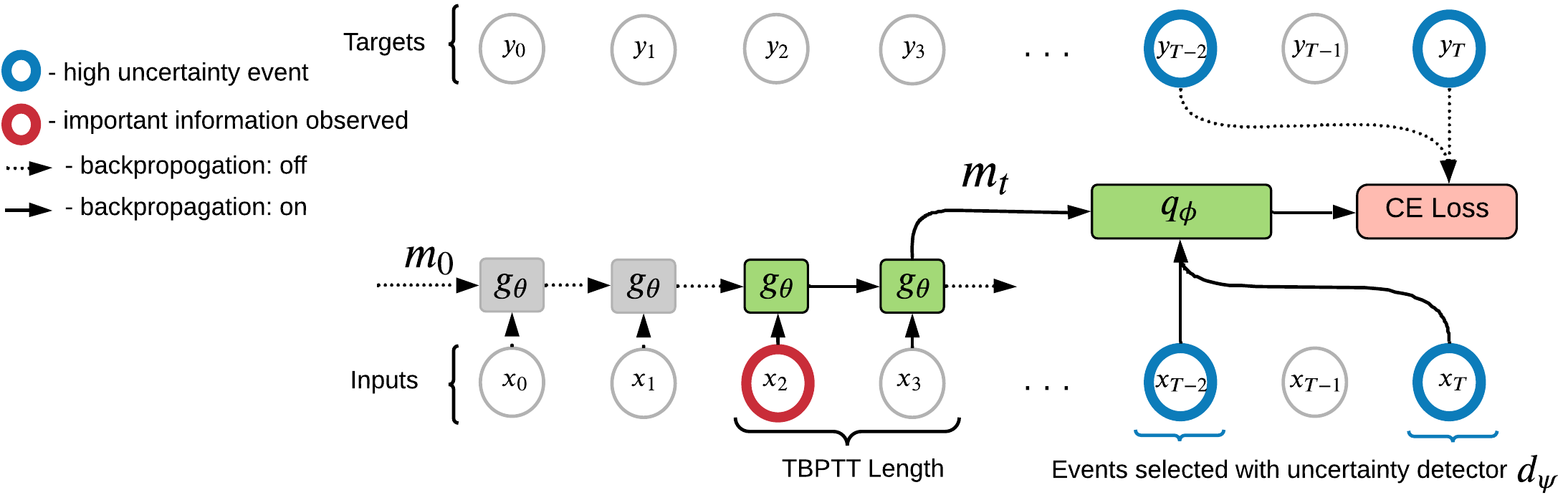}}
\caption{\small
\texttt{MemUP} gradient update. The recurrent memory $g_\theta$ processes a sequence. The number of prediction targets $K=2$ and TBPTT length $r=2$. Blue circles denote states with the highest uncertainty estimates $d_{\psi}(y_k|x_k)$. Here $x_2$ marked by red circle contains information that can help to predict outcomes $y_{T-2}$ and $y_T$. At the end of RNN TBPTT rollout these states are selected to form a set $U_3 = \{T-2, T\}$.  Then, $m_3$, $x_{T-2}$, $y_{T-2}$, $x_{T}$, $y_{T}$  are used to compute CE-Loss according to eq. \ref{eq:surrogate_objective}. While states 2 and T could be separated by many millions steps, \texttt{MemUP} allows memory to capture the utility of information from $x_2$ by propagating gradients only through  the sequence of $K+r=4$ elements at a time.
}
\label{fig:rnn_mem_train}
\end{center}
\vskip -0.2in
\end{figure*}

\textbf{Selecting states with high uncertainty.} Given uncertainty estimates $\{s_i\}^T_{i=t}$ for each future element in the sequence, we use softmax sampling $p(k) = e^{s_k/\tau}/\sum^T_{i=t} (e^{s_i/ \tau})$ without replacement with  $\tau = 0.02$. To implement sampling without replacement Gumbel-Max Trick \cite{gumbel_max_review_2021} is used. 
A scheme illustrating a single gradient update is shown in Figure \ref{fig:rnn_mem_train}.


\section{Related work}
\label{sec:related_works}

The problem of learning long-term dependencies has long been studied in both supervised learning~\cite{ rnn_tutorial_jaeger2002, rnn_gradient_hochreiter2001} and reinforcement learning domains~\cite{rl_with_memory_bakker2001}.
However, most of the research related to recurrent networks architectures is aimed at solving the problems of exploding and vanishing gradients~\cite{lstm_hochreiter1997, ntm_graves2014, dnc_graves2016, srnn}. 
The common trend is that many of the existing solutions to deal with vanishing gradients simultaneously increase the cost associated with the calculation of these gradients~\cite{ntm_graves2014, dnc_graves2016}.
For example, Transformers\cite{transformers_vaswani2017} generally require to store $O(N^2)$ of intermediate results for training on sequences of length $N$.
The recently proposed Linformer~\cite{linformer_wang2020} and BigBird~\cite{big_bird_zaheer2020} architectures allow training Transformers with linear complexity in space. 
However, most of these new Transformers have not yet found their way in the reinforcement learning setting.

In the field of deep reinforcement learning, the dominant approaches to implementation of memory are based on recurrent neural networks in combination with modern reinforcement learning algorithms~\cite{a3c_mnih, drqn_hausknecht2015, fun_hrl_vezhnevets2017, hide_and_seek_openai2020}. There are many studies that propose memory architectures for specific features of the reinforcement learning problems~\cite{r2d2_kapturowski2018, relational_rnn_santoro2018}. In the paper by Parisotto and Salakhutdinov~\cite{neural_map_parisotto2018}, a specifically modified \texttt{DNC}~\cite{dnc_graves2016} architecture learned long-term dependencies of several hundred steps in the setting of 2D and 3D navigation tasks. The downside of this architecture is that the agent needs access to the information about its absolute or relative location in a 2D/3D environment. 
In another work, Ha and Schmidhuber~\cite{world_models_ha2018} used a complex procedure with pre-training of memory and state embeddings. Their overall learning procedure is similar to ours. However, their memory/world model is trained to make predictions about the next step observations only, while we train memory to make predictions about events that can be arbitrarily far in the future.

Wayne at al.~\cite{merlin_wayne2018} demonstrated the state of the art results with the MERLIN algorithm. MERLIN's memory kept observation embeddings trained with a complex variational autoencoder in a \texttt{DNC}-like~\cite{dnc_graves2016} list. The memory module was also trained separately from the policy. 
Hung et al.~\cite{transporting_value_hung2019} followed up on the MERLIN architecture's success and used the soft attention mechanism over all past embeddings to encourage exploratory actions in the POMDP environment. Their experiments showed models learning temporal dependencies with a length of 500+ steps.
In contrast to our work, both of these solutions still require processing an arbitrarily long sequence of state embeddings by the policy networks at each step. 

The direct attempts in applying Transformer architecture in reinforcement learning setting resulted in conclusion that Transformers are too unstable to work properly in RL~\cite{snail_mishra2018}. Parisotto et al.~\cite{stabilizing_transformer_2019} were able to overcome the instability of Transformers, showing state of the art result in the memory dependent 3D tasks. 
Another recent work, by Loynd et al.~\cite{working_mem_graphs_loynd2020}, has also applied Transformers to the POMDP problems, but the WMG architecture relies on availability of factored observations to the agent which imposes additional requirements on the environment.
Switching focus from Transformers and complex architectures, Beck et al.~\cite{amrl_beck_2020} proposed \texttt{AMRL} architecture: a simple modification to the \texttt{LSTM} architecture to combat the problem of learning long-term dependencies in environments with a significant amount of observational noise.

\begin{figure*}[b!]
\vskip -0.2in
\begin{center}
\centerline{\includegraphics[width=1.0\textwidth]{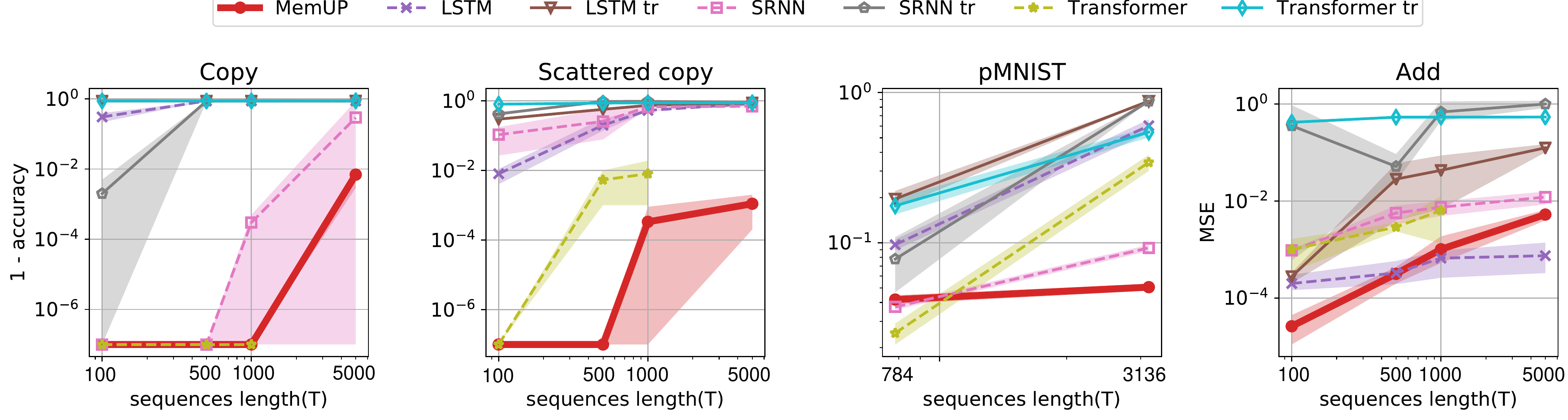}}
\caption{
Final performance on supervised learning tasks. X-Axis shows task scaling, while Y-axis correspond to metric score at the end of training. Metrics: \texttt{Inverted Accuracy} (1. - Accuracy) in Copy, Scattered copy and pMNIST tasks, \texttt{MSE} in Add task. All curves are averaged over 3 runs. 
}
\label{fig:sl_results}
\end{center}
\vskip -0.2in
\end{figure*}

\section{Supervised Learning Experiments}
\label{sec:sl_experiments}

For evaluation and comparison of our method we use four tasks: Copy \cite{copy_task}, Scattered copy, Add \cite{lstm_hochreiter1997} and permuted sequential MNIST (pMNIST) \cite{pmnist}. All these tasks are benchmarks that are used for testing models with long-term memory. In the original Copy task a sequence of length $l$ ($l=10$ in out experiments) should be copied after a $T-l$ steps ($T \in \{120, 520, 1020, 5020\}$ in our experiments), and in Scattered copy task a model has to make predictions in locations inside range $[l, T-1]$ that are chosen at random. We add this task in order to make  detection of  high-uncertainty locations harder.  For Copy, Scattered copy and Add tasks train and test datasets have sizes $10$K and $1$K, for pMNIST $60$K and $10$K. For detailed tasks description see Section \ref{sec:SL_tasks}.

In these experiments predictor $q_\phi$ consists of a three-layered MLP and a recurrent input encoder. The encoder $\rnnembed$ consists of a single fully-connected layer followed by two \texttt{LSTM}-layers with 128 hidden units and dropout probability $0.1$. $\rnnembed$ is used to encode bigger local context than original inputs corresponding to indices in the $U_t$ set (it is also possible to use a feedforward network for this task). Memory $g_\theta$ has the same architecture as $\rnnembed$ with separate weights. 

We do not train a separate detector $d_\psi$. Uncertainty at each step is estimated by prediction error from the predictor module $q_\phi$. For more information you can look at our implementation ( see section \ref{sec:code}).


In the Add Task the model is trained with MSE loss, which is a special case of CE loss under the model-assumption that target distribution is a Gaussian with unit variance. 
Both memory and encoder are trained with the same Truncated BPTT length. Truncation length equals $10$ in all tasks with sequence length $<1000$, $20$ in tasks with length $\geq1000$ and $30$, $60$ in pMNIST $784$, $3136$ respectively.



We compare \texttt{MemUP} with \texttt{LSTM} \cite{lstm_hochreiter1997} and \texttt{Transformer} \cite{transformers_vaswani2017} models from \texttt{pytorch} library, as well as a recurrent network \texttt{SRNN} \cite{srnn} which is designed for training on long sequences and has reduced  saturation of gradients. \texttt{LSTM*} and \texttt{SRNN*} have the same architecture as $\rnnembed$ with 3 layered MLP on top. In the case of \texttt{SRNN} baselines, we have replaced all \texttt{LSTM} layers with \texttt{SRNN} layers. The \texttt{Transformer} model consists of a single FC layer followed by $4$ self-attention layers and the same 3 layered MLP on top.
Baselines were trained in two modes. In the first mode (columns with Transformer, \texttt{SRNN}, \texttt{LSTM}) all elements of input sequence of length $T$ were fed into a model simultaneously, so  gradients propagate through the whole sequence during the training. We don't have results for  Transformer on sequences of 5000+ steps since it didn't fit into our GPU RAM. In the second mode (Transformer Tr., \texttt{SRNN} Tr., \texttt{LSTM} Tr.) BPTT rollout for recurrent architectures and attention window for Transformer were truncated to process the same number of steps as \texttt{MemUP}~($r+K$). 

Results in Figure \ref{fig:sl_results}
show that only \texttt{MemUP} allows to achieve a good quality of prediction in all tasks. 
In truncated BPTT setting (solid lines) MemUP is significantly better than \texttt{LSTM tr}, \texttt{SRNN tr} or \texttt{Transformer tr} due to \texttt{MemUP} ability to learn temporal dependencies longer that the length of BPTT rollout. 
When full information about the sequence is available to \texttt{LSTM}, \texttt{SRNN} and \texttt{Transformer} (dashed lines) but not \texttt{MemUP}, the latter still shows the best results for longest versions of \texttt{Copy}, \texttt{Scattered Copy}, \texttt{pMnist}, and has better overall performance.
Moreover, when compared with Full BPTT baselines, \texttt{MemUP} requires to store much less intermediate computations ($10$-$250$ times) due to usage of short Truncated BPTT rollouts. 

\section{Reinforcement Learning Experiments}
\label{sec:rl_experiments}
\begin{figure}[b!]
\begin{center}
\centerline{\includegraphics[width=0.7\columnwidth]{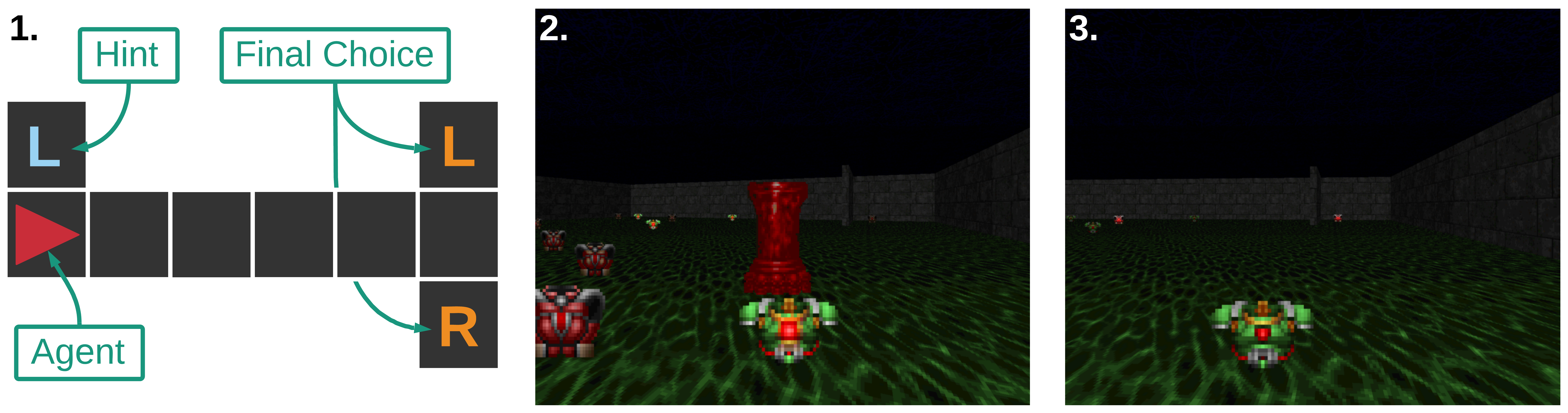}}
\caption{
\small{
Environments for testing long-term time dependencies. \textbf{1.} \textit{T-Maze environment.} The agent should reach the T-shaped junction and choose one of the arms (L or R). A hint about what arm to choose is provided at the very beginning. \textbf{2.} and \textbf{3.} \textit{Vizdoom-Two-Colors environment}. The agent is in the room and constantly loses health. To replenish his health and recieve a reward the agent needs to collect items of the same color as the column. The color of the column is chosen randomly from red or green options at the beginning of the episode. After 45 steps from the beginning the column disappears. The episode lasts for a maximum of 1050 steps.
}
}
\label{fig:all_envs}
\end{center}
\vskip -0.4in
\end{figure}

Performance of the \texttt{MemUP} algorithm is studied in two reinforcement learning tasks.
The first task is classic T-Maze environment where the main difficulty is a long-term dependency between a hint at the starting position and location of a reward at the exit (see Figure \ref{fig:all_envs}.1). The main advantage of this problem is that it allows to test the agent's memory mechanisms in isolation because of primitive policy and the observation space. We base our experiments on the noisy version introduced by~\cite{amrl_beck_2020}.

The second task is the color dependent object collection in  the Doom environment ~\cite{vizdoom_kempka2016, fps_vizdoom_lample2017} which requires long-term memory in combination with a more complex policy and rich observation space. In the task introduced by~\cite{drl_on_budget_2019}, the agent is placed in a room filled with acid (see Figure \ref{fig:all_envs}.2.) and constantly loses health. Objects of green and red colors are scattered throughout the environment. Object of one color replenish the agent's health and give a +1 reward, while others take away health and give a -1 reward. The correspondence between effects and colors is determined randomly at the beginning of each episode. This information is conveyed by a column whose color matches the color of health replenishing items. In our version the column disappear after first 45 steps (see Figure \ref{fig:all_envs}.3). The episode ends when the agent's health drops to zero or after 1050 time steps.

The most straightforward extension of the MemUP algorithm to a reinforcement learning problem is to use memory state $m_t$ as an additional input to an RL agent.
There are several possible solutions to combine MemUp's training with agent's policy training:  (1) pre train MemUP and then train an agent, (2) use alternating memory and agent training phases, train memory and agent in parallel, (3) combining the MemUP and RL agent into a single into an end-to-end network. 
In these experiments, we test the simplest version:  train MemUP on trajectories generated with a frozen policy, then train an agent with a frozen MemUP memory. The second version is also implemented in the code (see Section \ref{sec:code}).

Inputs $x_t$ are formed from observations $o_t$ and previous action $a_{t-1}$. This time Observations are encoded using convolutional networks or fully connected networks. 
We use discounted future returns $R_t = \sum_{i = t}^T \gamma ^ {i-t} r_i$. as targets $y_{t}$. In the experiments, \texttt{MemUP} is compared with the following baselines:


 \textbf{PPO-LSTM} is a recurrent version of \texttt{PPO}~\cite{ ppo_schulman2017}  with a single \texttt{LSTM}-layer~\cite{ lstm_hochreiter1997}. 
 We use \texttt{PPO}-\texttt{LSTM} implementation from the RLPyt library~\cite{ rlpyt_stooke2019}.
    
 \textbf{IMPALA-ST} is an IMPALA~\cite{impala_espeholt2018} agent using Stabilized Transformer\cite{transformers_vaswani2017} architecture. Stabilized Transformer was presented by Parisotto et al.~\cite{stabilizing_transformer_2019}. For experiments, we use the only available open implementation of this algorithm ~\cite{adaptive_transformers_rl_kumar2020}. 
    
\textbf{AMRL} is proposed by~\cite{amrl_beck_2020} for the Noisy T-Maze Task. In the original study \texttt{AMRL} was compared with many different memory architectures including \texttt{DNC}, multi-layered \texttt{LSTM} and outperformed all of them on T-Maze-LN-100 task as well as other memory experiments. \texttt{AMRL} is similar to \texttt{PPO}-\texttt{LSTM} baseline, but extends \texttt{LSTM} with \texttt{AMRL} Layer. 

In all experiments we use the same encoder networks for observation embeddings for all baselines and \texttt{MemUP}. Additionally the size of an \texttt{LSTM}-layer is the same for the memory module $g_\theta$, \texttt{PPO}-\texttt{LSTM} and \texttt{PPO}-\texttt{AMRL} baselines.
\begin{figure*}
\vskip 0.2in
    \subfloat[\label{fig:t_maze_1k_results_agent}]{\includegraphics[width=0.48\columnwidth]{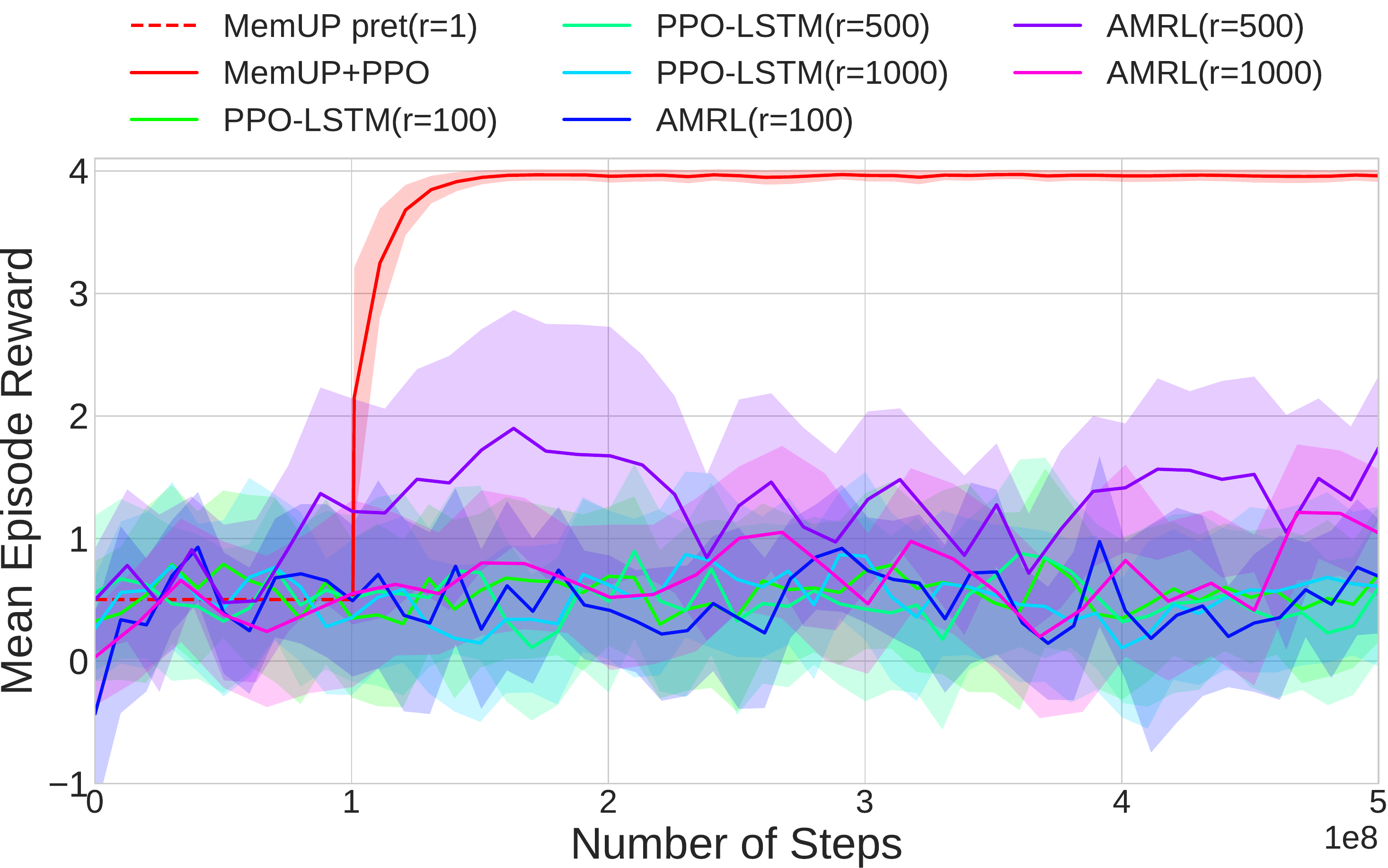}}
\hfill
    \subfloat[\label{fig:vizdoom_results}]{\includegraphics[width=0.48\columnwidth]{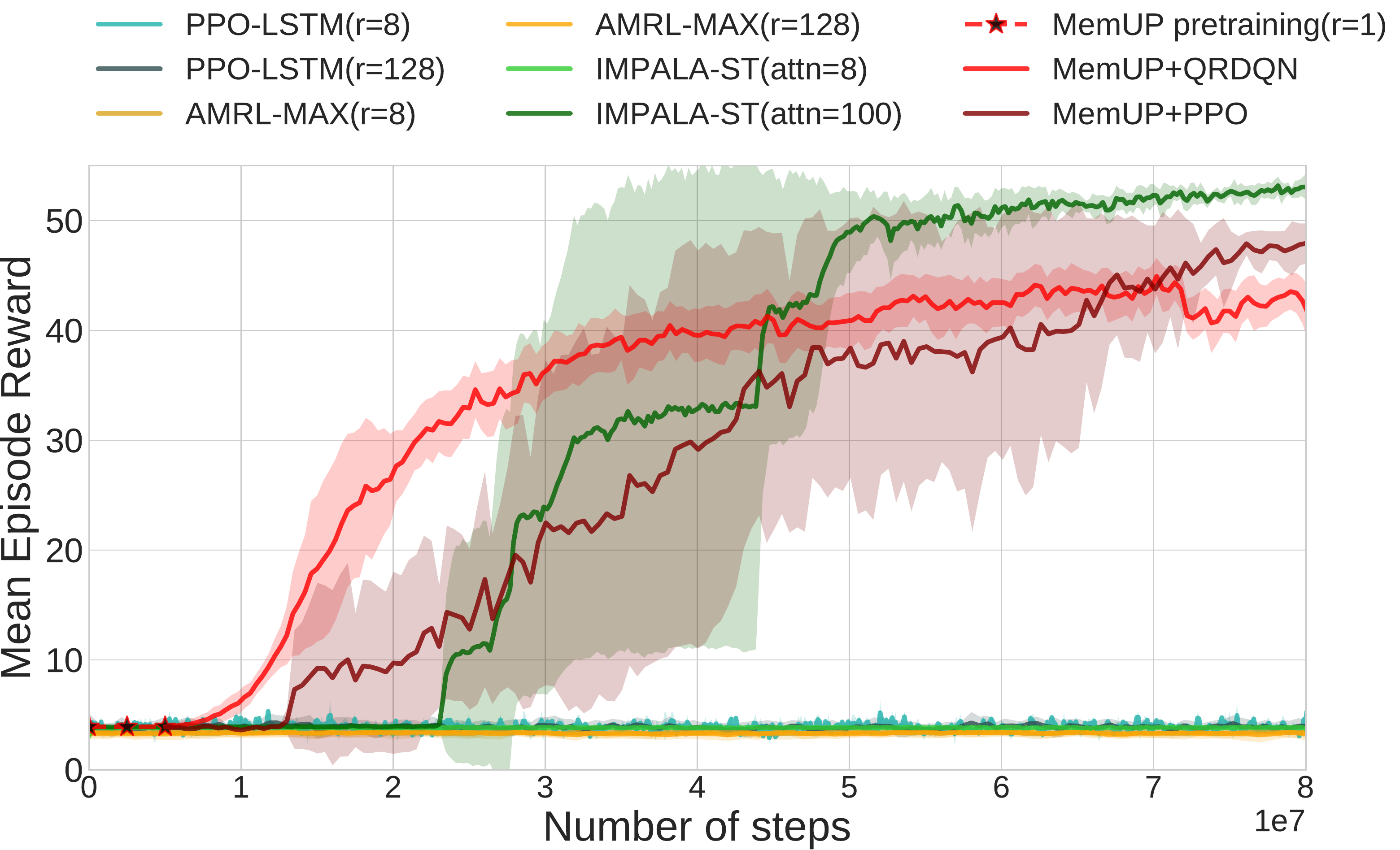}}    
\caption{\small{
\textbf{a)} Learning curves for \texttt{MemUP} and baselines in T-Maze-LNR-1000. 
\textbf{b)} Learning plots in the Vizdoom-Two-Colors environment for all agents. All curves are averaged over 3 runs. Translucent areas around each curve show the standard deviation computed over all runs. The memory pre-training phase of \texttt{MemUP} agent is marked with the dashed line. TBPTT length/attention span for each baseline is specified in the parenthesis.}}
\label{fig:rl_results}
\vskip -0.25in
\end{figure*}
Results for the T-Maze-LNR-1000 (10 times longer dependency than in~\cite{amrl_beck_2020}) environment are presented in Figure \ref{fig:t_maze_1k_results_agent}. The designation T-Maze-LNR-1000 means that the maze length (and corresponding tempral dependency) in each episode can be from 1000 to 1009 steps. 
Several rollout lengths are tested for RNN-based baselines. We could not successfully train a single \texttt{PPO}-\texttt{LSTM} or \texttt{AMRL} agent even with TBPTT rollout of $r=1000$\footnote{We tested our AMRL implementation on T-Maze-LNR-100 with full rollout and get results aligning with the original paper. But the algorithm did not withstand 10 times increase in episode length and truncation constraints}. 
During MemUP pre-training phase, episodes are generated with a random policy. We use the discounted future reward with $\gamma = 0$ as a prediction target.
A policy training for the \texttt{MemUP} agent starts after the memory pre-training phase is completed. The time spent in the pre-training phase is marked with a dashed line in all figures.
As shown in Figure
 \ref{fig:t_maze_1k_results_agent} the agent learns almost instantly, given a memory pre-trained with our method. For the \texttt{MemUP} agent we train all components including memory module and uncertainty detector for each individual run from scratch. Additionally, Section \ref{sec:long_tmaze}
 shows that \texttt{MemUp} can be trained to solve T-maze-LNR-20000.


Results for the Vizdoom-Two-Colors environment are shown in Figure \ref{fig:vizdoom_results}. 
IMPALA-ST(attn=8) with a short attention span could not learn any reasonable policy. On the other hand, IMPALA-ST(attn=100) solves the problem and reliably survive in the environment for 1050 steps. 
Both \texttt{PPO}-\texttt{LSTM} and \texttt{AMRL} baselines with TBPTT rollouts of 128 and 8 steps could not learn to survive longer than a random agent. While the rollout of 128 steps is potentially enough to remember the color of the column or a previously collected items, \texttt{PPO}-\texttt{LSTM} and \texttt{AMRL} agents are unable to utilize long-term information from the past. 

For \texttt{MemUP} we use the discounted future reward with $\gamma = 0.8$ as a prediction target.
The memory module $g_\theta$ is trained with the rollout length of 1 step.  In Vizdoom-Two-Colors we set number of targets $K=3$. Thus, to train the memory $g_\theta$ we use 4 separate observations per episode for one gradient update. 
In ViZDoom, we train \texttt{PPO} (\texttt{MemUP}+\texttt{PPO}) and \texttt{QRDQN} (\texttt{MemUP}+\texttt{QRDQN}) agents with MemUP pretraining.


Both \texttt{MemUP}+\texttt{PPO} and \texttt{MemUP}+\texttt{QRDQN} learn to survive in the environement for full 1050 steps by collecting healing items. Though the final performance of IMPALA-ST(attn = 100) is slightly better. We speculate that this is
due to the fact that the pre-trained memory \textbf{M} has learned to store mostly information about the color of the column, while IMPALA-ST (attn=100) has slightly better spatial awareness. 
On the other hand, \texttt{MemUP} use significantly fewer resources in terms of the size of processed sequences during training/pre-training and evaluation.
As well, \texttt{MemUP} significantly outperforms baselines with comparable resource requirements, like IMPALA-ST (attn=8), \texttt{PPO}-\texttt{LSTM}(r=8) and \texttt{AMRL}-\texttt{MAX}(r=8).


\section{Sensitivity to the Noisy-TV problem}
\label{sec:noisy_tv_problem_main}

\begin{figure*}[b!]
\begin{center}
\includegraphics[width=0.4\textwidth]{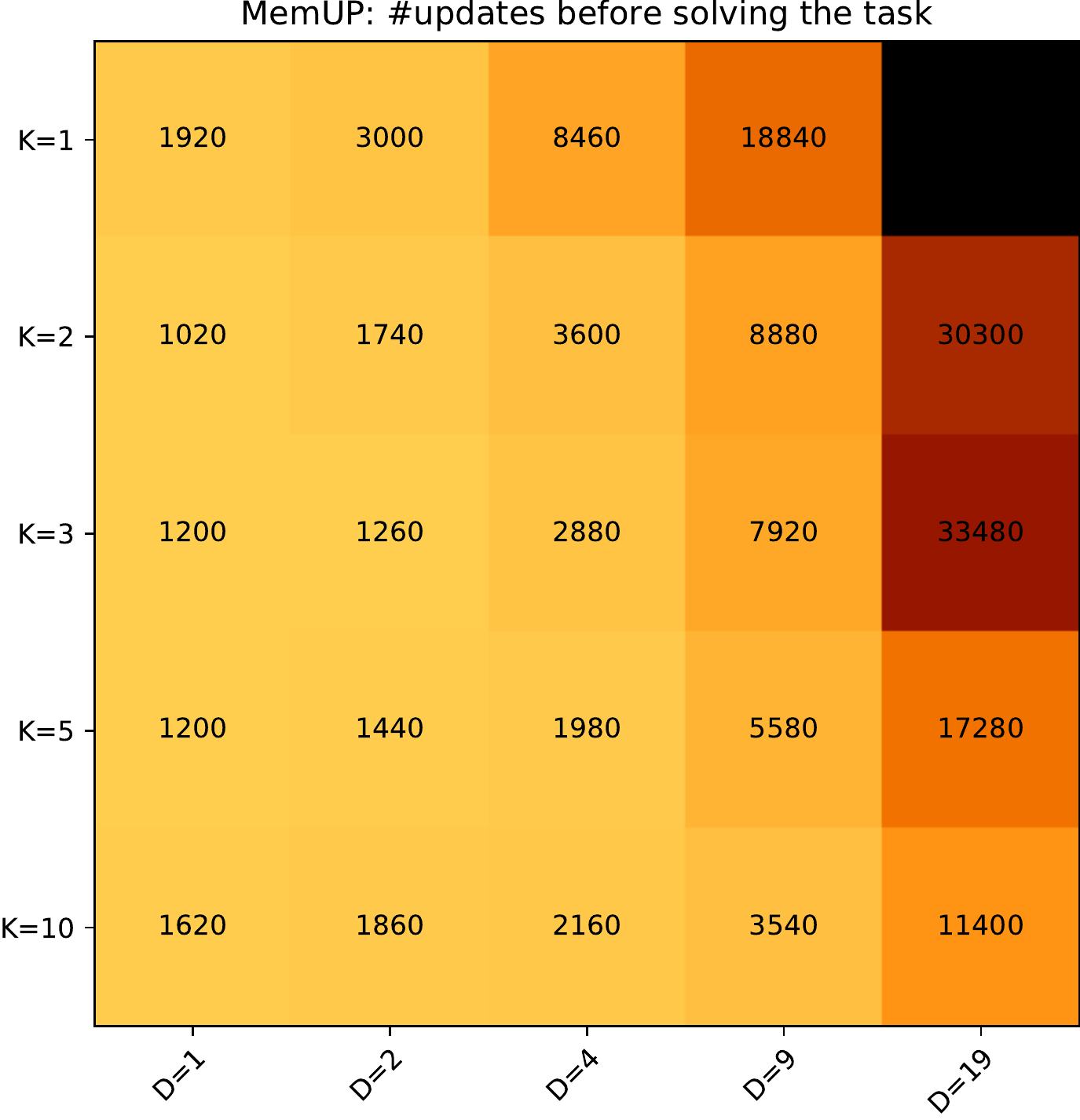}
\caption{
Each cell shows the mean number of updates required for the memory to correctly predict (achieving $0.1$ RMSE) the reward after the decision point connected with the temporal dependency. The x-axis shows the number of unpredictable decision points in the environment. The Y-axis shows number of events predicted by \texttt{MemUP}. Darker colors means slower solution.
}
\label{fig:noisy-tv-heatmap}
\end{center}
\end{figure*}

In section \ref{sec:idea}  we discussed  conditions on true uncertainty of targets $y_t$ given a perfect memory  $m^{*}_t$. The goal was to set conditions under which we can make a good guess about relative values of potential memory utility $I(y_t|m^{*}_t, c_t)$  by estimating local entropy $H(y_t|c_t)$. 
However, this may not be the case.
For example, some targets $y_t$ may be completely unpredictable and no information from the past can reduce the entropy for the distribution of $y_t$: $H(y_t|x_t) \sim H(y_t|m^{*}_t, x_t)$. Then the local entropy is large, while potential utility of memory for $y_t$ predictions is close to zero. Training memory to predict such targets is pointless and  prevents \texttt{MemUP} from focusing on events for which training long-term memory is essential. 
The problem of over-emphasis on unpredictable surprising events is called Noisy-TV problem\cite{rnd_burda2018}. Noisy-TV problem is often encountered by algorithms using curiocity-based exploration in reinforcement learning\cite{large_curiosity_study_burda2018}.
To test robustness of \texttt{MemUp} to the Noisy-TV problem, we have modified T-maze-LNR-100 environment. 
\paragraph{Noisy T-maze With Distractors.} In this version, agent can receive +4 or -3 rewards in D+1 decision points. Their location in the corridor is chosen randomly at a distance of at least 50 steps from the hint. All of them can be detected when $o_t[1] != 0.$ element in observation vector $o_t$. Decision points are distinguishable from each other. Each decision point has a unique value of $o_t[1]$. Only in one decision point, the next reward depends on the agent's action and the value of the hint ($o_t[1] = 1.$ in this case ). In other D decision points, the next rewards are completely random ( chosen with a probability of 0.5 ).
Thus, Noisy T-maze With Distractors generate sequences with one long-term dependency and D events acting as Noisy-TV distractors.

We conducted 25 experiments with 6 independent runs in each (150 runs in total). The differences in the experiments come down to two parameters: number of distracting events D (X-axis in Fig. \ref{fig:noisy-tv-heatmap}) and number of events sampled for prediction K (Y-axis in Fig. \ref{fig:noisy-tv-heatmap}).
 More distracting events make the task of learning the temporal dependency harder. Increasing the size of $\mathcal{U}_t$ set( by changing $K$) improves the chance of selecting a future event that is a part of temporal dependency. On the other hand, with an increase in $\mathcal{U}_t$, the resource cost for memory training also increase. The TBPTT rollout length in these experiments is 1.

Results in Fig. \ref{fig:noisy-tv-heatmap} show that with an increase in the number of distractions, memory learning speed decreases. However, increasing the set $\mathcal{U}_t$ allows to alleviate this problem. For the experiment (D=19, K=1), none of the runs solved the task in the first 45000 updates. In experiments (D=19, K=2) and (D=19, K=3) only 4 out of 5 runs solved the task. The results show that the agent can learn the time dependency fairly quickly even if the uncertainty detector select noisy events $80\%$ of the time.

\section{Ablation study}
\label{sec:ablation_study}
To study an effectiveness of our core idea, i.e. training memory by predicting long-term future events with high uncertainty, we compared \texttt{MemUP} with baselines that exclude core \texttt{MemUP} features while sharing the same neural architecture. 

In these experiments, we carefully consider the process of learning memory (without further training the RL agent) on the T-Maze-LNR problems.
\begin{figure*}[b!]
\begin{center}
\subfloat[\label{fig:ablation_err_per_rollout_tm100}]{\includegraphics[width=0.45\columnwidth]{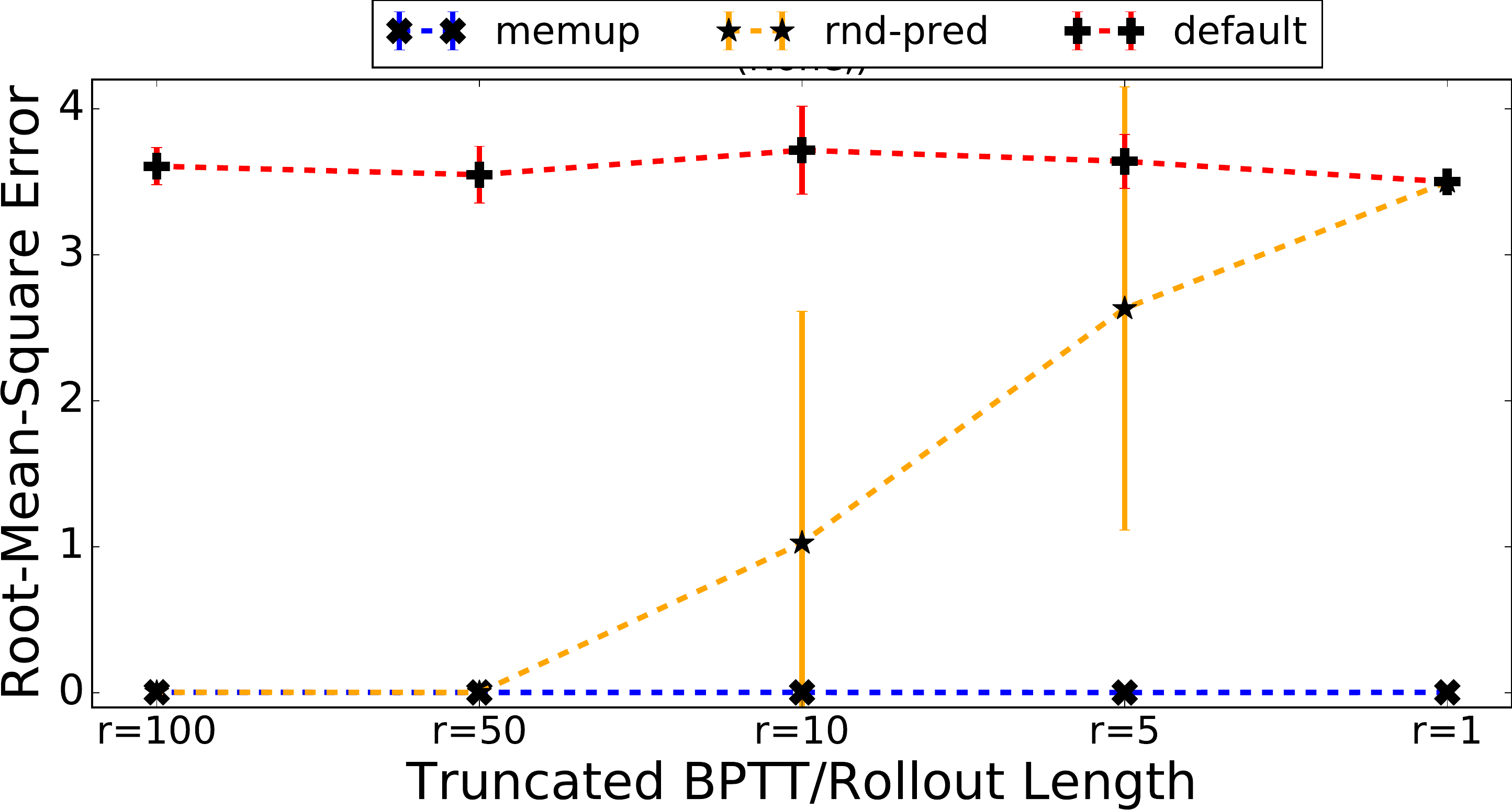}} 
\hfill
\subfloat[\label{fig:ablation_err_per_rollout_tm1k}]{\includegraphics[width=0.45\columnwidth]{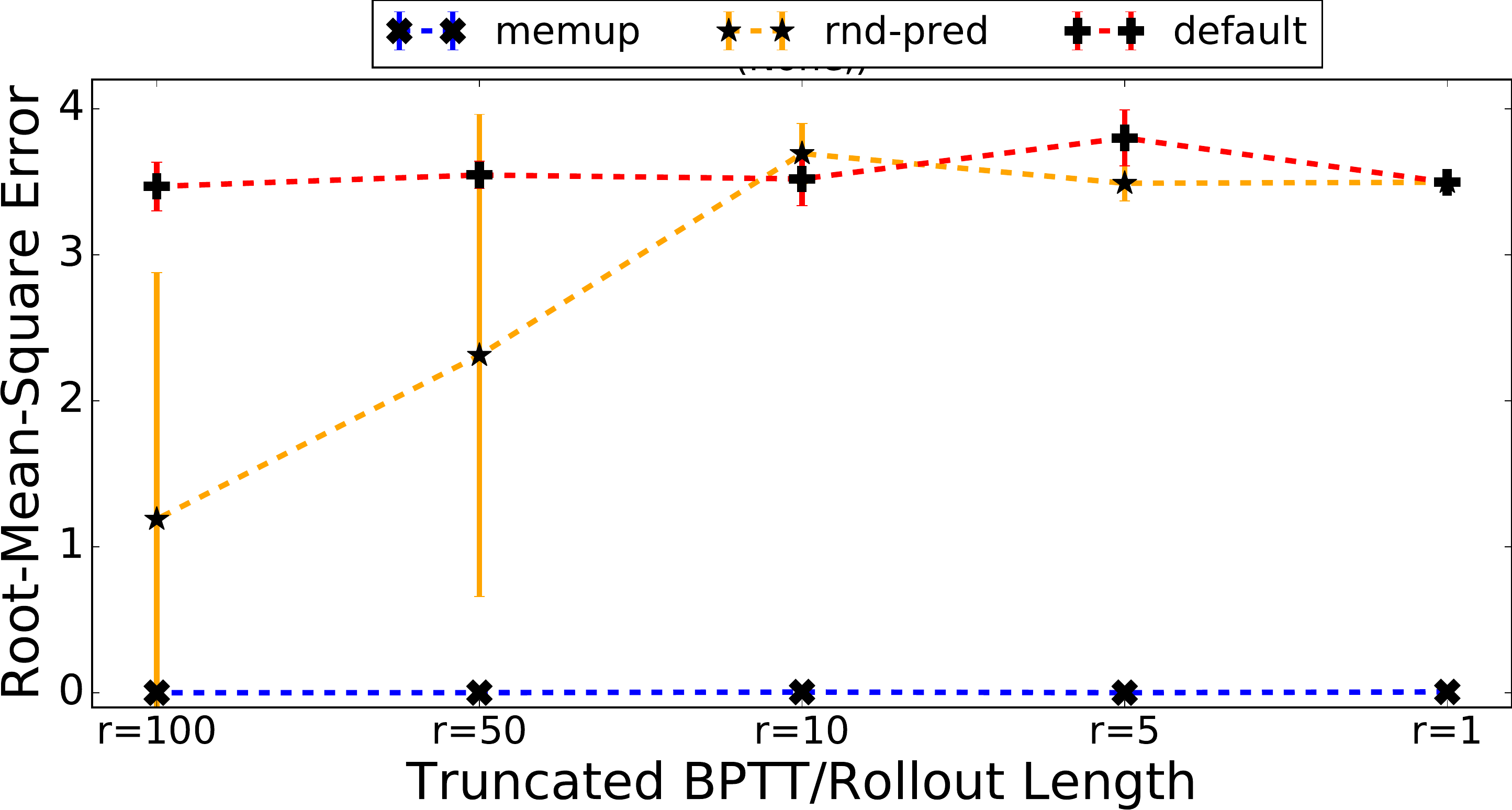}}  

\caption{
Averaged over 6 runs prediction error at the end of training for each ablation baseline with respective TBPTT length. Vertical bars show standard deviation computed over 6 runs. \textbf{a)} T-Maze-LNR-100. \textbf{b)} T-Maze-LNR-1000. 
}
\label{fig:ablation_error_per_rollout}
\end{center}
\end{figure*}
The following memory ablation baselines were studied:

\textbf{MemUP:} Proposed in this paper. See section ~\ref{sec:idea}. 

\textbf{Rnd-Pred:} Same as \texttt{MemUP}, but steps for future outcome prediction are selected randomly and uniformly among all future timesteps. 
This randomization verify whether predicting steps with high uncertainty helps \texttt{MemUP} to overcome the problem of learning long-term dependencies with small BPTT rollouts.

\textbf{Default:} Same as \texttt{MemUP}, but it is trained to predict return $R_t$ at each step $t$, as oposed to an arbitrarily distant future events as in \texttt{MemUP} and \textit{Rnd-Pred}. Thus, \textit{Default} is the same as the \texttt{LSTM} baseline in the Supervised Learing Experiments (see \ref{sec:sl_experiments}). But in this study we test this baseline on all intermediate Truncated BPTT lengths. 
   
All versions have the same architecture, and hyperparametes. They were trained in T-Maze-LNR-100 and T-Maze-LNR-1000 problems.  Episodes were generated by a random strategy. All versions are trained with 5 different Truncated BPTT lengths: 1, 5, 10, 50, 100. Testing metric is a root-mean-squared error(RMSE) between the models' prediction and the actual return at the final step of the episode. 
The evaluation was conducted on the 100 separately recorded episodes.

Figure \ref{fig:ablation_error_per_rollout} shows the results of the experiments. On the x-axis, we have specified the length of the Truncated BPTT in decreasing order, i.e. in the order of increasing task complexity.  For every rollout length there is a corresponding absolute value of error between predictions of the model and targets at the end of training ($3e4$ updates for T-Maze-LNR-100 and $3e5$ for T-Maze-LNR-1000). Smaller values represent better predictions.

The \textit{Default} modification is unable to learn a long-time dependency exceeding the length of the tested rollouts. This is the same as the results of the \texttt{PPO}-\texttt{LSTM} agent on the T-Maze-LNR Reinforcement Learining experiments. 

The \textit{Rnd-Pred} modification solves the task with rollouts 100 and 50 in T-Maze of length 100, but for shorter rollouts and for longer T-Maze-LNR-1000 results deteriorate significantly. In the majority of runs in T-Maze-LNR-1000, the memory has not been learned.  The relative success of \textit{Rnd-Pred} is expected because random  selection of prediction target still allows to sample a useful future event with some small probability.\footnote{For example, for a rollout of 50 steps, the chance to select the correct target step is 0.5} Thus, random sampling of prediction targets allows to learn casual dependency that is 2 times longer than TBPTT, but already struggles at ratio of 10. 

On the other hand, the \texttt{MemUP} is able to solve the problem regardless of the rollout length, i.e. \texttt{MemUP} demonstrated an ability to learn casual dependencies that 1000 times longer than its TBPTT length in this setting. In this study \texttt{MemUP} is successfully trained with TBPTT of length 1, in other words, without using backpropagation through time at all.

\section{Conclusion}
\label{sec:conclusion}

In this paper, we proposed a new method for training long-term memory. The main idea is to train a memory network to predict future outcomes of high uncertainty and skip all others. Predicting a small number of arbitrarily distant future outcomes substantially saves computational resources required to backpropagate gradients. At the same time, the emphasis on high uncertainty outcomes allows not to miss long-term dependencies in the task.

Experimental results show that our training algorithm allows to learn temporal dependencies significantly longer than the number of steps processed for a single gradient update.
None of the baseline architectures trained in a classical way demonstrate such ability. All these baselines with comparable results require to store at least 200 times more intermediate calculations for the Add, Copy, Scattered Copy, pMNIST tasks as well as at least 500 times more for the T-maze, and 625 times more for the Vizdoom-Two-Colors (Stabilized Transformer baseline has quadratic complexity while processing 25 times more steps).

Even using fewer resources, MemUP outperforms non-truncated baselines that has simultaneous access to all elements of a sequence (including Transformer) on pMNIST(3136), Copy and Scattered Copy tasks. On T-maze and Vizdoom-Two-Colors tasks, MemUP is better than all recurrent baselines. Another advantage is that  MemUP can be combined with any recurrent architecture and applied both to supervised as well reinforcement learning settings. 

We believe that the MemUP algorithm applied to recurrent networks has shown promising results, and the main idea demonstrates exciting avenues for future research.

\bibliography{mem_up}

\newpage
\appendix

\section{Mutual Information Maximization}
\label{sec:optimization}

Estimating mutual information $I( y_k,m^\theta_t|x_k)$ can be difficult task from a computational point of view. Therefore, to make the task more trackable we use a lower bound for mutual information derived by \cite{im_agakov2004}. Adapting it for the \texttt{MemUP} case, we get the following inequality:
\begin{align}
    \label{eq:information_lower_bound}
    I(y_k, m^{\theta}_t | x_k) &= H(y_k|x_k) - H(y_k|m^{\theta}_t, x_k) \nonumber  \\ 
 &\ge H(y_k|x_k) - CE(p(y_k|m^{\theta}_t, x_k), q_{\phi}(y_k|m^{\theta}_t, x_k)), 
\end{align}
where $m^{\theta}_t$ is a shortcut notation denoting dependence of the memory state $m_t$ on the parameters of memory network $g_{\theta}$. The distribution $p$ is the true conditional distribution of future outcomes $y_k$, which is unknown to us, but we can sample from it using training data. The distribution $q_\phi$ is approximated by the predictor network with parameters $\phi$. Inequality \ref{eq:information_lower_bound} follows from the relation between KL-divergence, cross-entropy (CE) and entropy: $D_{KL}(p || q_{\phi}) =CE(p,q_{\phi}) - H(p)$. Using non-negativity property of  KL-divergence leads to $CE(p,q) \ge H(p)$. The bound becomes exact when $q_{\phi}(y_k|m^{\theta}_t, x_k)$ is equal to $p(y_k|m^{\theta}_t, x_k)$ 

Since $H(y_k|x_k)$ is independent from memory and predictor networks, maximizing the lower bound is the same as minimizing cross-entropy. Therefore, the memory $g_{\theta}$ and the predictor $q_\phi$ can be jointly trained to maximize mutual information by simply minimizing cross entropy loss:
\begin{align}
    \label{eq:lower_bound_objective}
    \min_{\theta, \phi} 
    \sum^{T}_{k \in \mathcal{U}_t}   CE(p(y_k|m^{\theta}_t, x_k), q_{\phi}(y_k|m^{\theta}_t, x_k))
    = 
    \min_{\theta, \phi} 
    \sum^{T}_{k \in \mathcal{U}_t} \mathbb{E}_{x_k, y_k}[-\log\, q_\phi (y_k| m^{\theta}_t, x_k)].
\end{align}
Specifically, the memory module $g_\theta$ is trained by backpropagating gradients directly through the predictor network $q_\phi$. 


\section{Supervised Learning Tasks}
\label{sec:SL_tasks}
For accuracy evaluation we involve four tasks: Copy, Scattered copy, Add and permuted MNIST (pMNIST). All these tasks are benchmarks that are used for testing models with long-term memory.

\paragraph{Copy task.} This is an aligned sequence to sequence classification problem ($X \to Y$). Denote by $T$ length of $X$ and $Y$.  The first $10$ elements contain random uniform digits from set $\{2, \ldots, 9\}$ that model has to copy and reproduce in the end of the sequence where $X_i = 1$, $i \in \{T-10, \ldots ,T-1\}$. In all other positions except the first $10$ and the last $10$ $X_i = 1$, $i \in \{10, \ldots ,T-11\}$. Correspondingly $Y_i = 0$, $i \in \{0, \ldots ,T-11\}$ except the last $10$ where $Y_{T-10+i} = X_{i}$, $i \in \{0, \ldots ,9\}$. We split this task into four different sub-tasks by the length where $T \in \{120, 520, 1020, 5020\}$. Train dataset contains $10$k sequences and test consists of $1$k sequences. 

\paragraph{Scattered  copy task.} It is similar to the previous one.
The only difference is that locations where one has to make predictions are choosen at random in range $[10, \ldots, T-1]$. 
In such locations $X_i = 1$ and $Y_i$ equals to some element from the first $10$ digits. We split this task into four different sub-tasks by the length where $T \in \{120, 520, 1020, 5020\}$. This task is more complex because it requires locations detection and their ordered number counting. 
As a metric in the previous two tasks we use negative log-likelihood (\texttt{NLL}). Train dataset contains $10$k sequences and test consists of $1$k sequences. 

\paragraph{Add task.}
This is an aligned sequence to sequence regression problem ($X \to Y$). Each element in $X$ is two dimensional. The first dimension contains a random uniform value $\in [0, 1]$ and second component is $0$ or $1$. There are only three ones per sequence that correspond to two summands and their sum. Values of $Y$ elements equal to zero except one element at the sum location. \texttt{MSE} metric was used here as a loss function and quality measure. Train dataset contains $10$k sequences and test consists of $1$k sequences.  We split this task into four different sub-tasks by the length where $T \in \{100, 500, 1000, 5000\}$.

\paragraph{Permuted sequential MNIST task (pMNIST).} The dataset is obtained from the ordinary MNIST that includes $60$k train images and $10$k test images. Each image is flattened to one dimensional vector. It yields vectors of size $784$ and we also add zero-value padding to obtain vectors of size $3136$. A random permutation changes order of the vector elements consistently. In training procedure we apply  \texttt{NLL} loss function and measure quality by classification accuracy in \%.    

\begin{table*}[t!]
\begin{small}
\caption{
\small{
Supervised Learning Results.
\textbf{ Metrics:}  \texttt{Accuracy} (classification accuracy in \%) in Copy, Scattered copy and pMNIST tasks, \texttt{MSE} in Add task. \textbf{Bold} font highlights the best score per task, * shows the best score among truncated methods.  
}
}
\label{table:supervised_experiments}

\vskip -0.1in
\begin{center}
\begin{tabular}{ | l | c | c | c | c | c | c |c| }
\hline
Task/Method       & MemUP    &  LSTM    & LSTM tr & SRNN    & SRNN tr & Transf. & Transf. tr \\ \hline \hline
Copy 120        & \textbf{100}*    & 69.3     & 12.7 & \textbf{100}   & 99.7  & \textbf{100} & 12.5  \\
Copy 520        & \textbf{100}*    & 12.4     & 13.0 & \textbf{100}   & 13.1  & \textbf{100} & 12.7 \\
Copy 1020       & \textbf{100}*    & 12.4     & 12.9 & 99.9          & 12.5  & \textbf{100} & 12.6 \\
Copy 5020       & \textbf{99.3}*    & 12.4     & 12.5 & 70.8          & 12.5  & out of mem & 12.5 \\
\hline
Scat. copy 120        & \textbf{100}*    & 99.2     & 70.4 & 89.4   & 57.8  & \textbf{100} & 20.3  \\
Scat. copy 520        & \textbf{100}*    & 80.1     & 43.4 & 74.5   & 5.0  & 99.5 & 14.3 \\
Scat. copy 1020       & \textbf{99.9}*  &  47.2     & 26.2 & 33.4   & 5.1  & 99.3 & 12.9 \\
Scat. copy 5020       & \textbf{99.9}*  &  16.8     & 12.9 & 30.3   & 12.5  & out of mem & 12.9 \\
\hline
Add 100           & \textbf{0.00003}* & 0.00019 & 0.00027 & 0.00095 & 0.356    & 0.00103  & 0.420   \\
Add 500           & \textbf{0.00031}* & 0.00032 & 0.02830 & 0.00565 & 0.516   &  0.00291  & 0.536    \\
Add 1000          &  0.00101* & \textbf{ 0.00066 } & 0.04294 & 0.00744 & 0.685    & 0.00638  & 0.537    \\
Add 5000          &  0.00526* & \textbf{ 0.00074 } & 0.12550 & 0.01206 & 1.000    & out of mem  & 0.546    \\
\hline
pMNIST 784       &  95.4*     & 89.5     & 79.85    & 96.43    & 95.33    & \textbf{97.1}     & 84.55    \\
pMNIST 3136      & \textbf{94.3}*    & 33.6     & 11.7     & 90.31    & 11.7     & 63.5     & 49.7 \\
\hline
\end{tabular}
\end{center}
\end{small}
\vskip -0.1in
\end{table*}

\begin{table*}[t!]
\begin{small}
\caption{
\small{
Memory size impact. We compute for MemUP method the dependence of \texttt{MSE} metric on LSTM layer's dimension in tasks  Add 500 and pMNIST 784. 
}
}
\label{table:memory_size}

\vskip -0.1in
\begin{center}
\begin{tabular}{ | l | c | c | c | c | }
\hline
Task/Memory size       & 128    &  256    &  512  &   1024  \\ \hline \hline
Add 500                &   0.00028    &  0.00018      &  0.00024    &   0.00056     \\
pMNIST 784             &  89.4  & 92.5    &  95.4 &   96.1  \\
\hline
\end{tabular}
\end{center}
\end{small}
\vskip -0.1in
\end{table*}

\section{Reinforcement Learning Tasks}

\label{sec:RL_tasks}
 \paragraph{Noisy T-Maze.} Namely, the agent starts at the very beginning of the central corridor next to the hint that it observes in the first step of the episode. The agent can only move forward along the corridor. At each step, the agent's observations $o_t$ are represented by a vector of length 3. The first element $o_t[0]$ contains value of the hint: +1 or -1 at the first step, 0 otherwise. The second element $o_t[1]$ is the indicator of reaching the intersection, which is 1 if the agent has reached the place of turn. The last element $o_t[2]$ does not carry information and is a random noise, it is +1 or -1 with equal probability. The agent receives the reward only at the end of the episode. It is +4 if the correct turn was chosen and -3 otherwise.

The length of the maze is defined as the number of steps between the moment the agent sees the hint and the moment the agent has to make a choice based on the value of the hint. The real maze length in experiments varies within 10 steps from episode to episode in order to decorrelate the observations of agents trained in parallel on multiple instances of environments (see \texttt{PPO} algorithms ~\cite{ppo_schulman2017}, A3C ~\cite{a3c_mnih}, IMPALA ~\cite{impala_espeholt2018} ). The naming of the environment indicates the minimum possible maze length, for example, the designation T-Maze-LNR-100 means that the maze length in each episode can be from 100 to 109 steps.

\paragraph{Vizdoom-Two-Colors.} Despite the fact that T-Maze environment allows simulating very long temporal dependencies, it is too simplistic.  To explore if \texttt{MemUP} augmentation can be scaled to much more complex tasks and environments it was tested in the Vizdoom-Two-Colors task introduced by Beeching et al.~\cite{drl_on_budget_2019}. In this task, the agent is placed in a room filled with acid (see Figure \ref{fig:all_envs}.2.) and constantly loses health. Objects of green and red colors are scattered throughout the environment. Items of one color replenish the agent's health and give a +1 reward, while others take away health and give a -1 reward. The correspondence between effects and items' colors is determined randomly at the beginning of each episode. This information is conveyed by a column whose color matches the color of health replenishing items. The episode ends when the agent's health drops to zero or after 1050 time steps. At each step agent receives a small "living reward" equal to +0.02.
Accordingly, the goal of the agent is to survive as long as possible in the environment by collecting items of the rewarding color.

To solve this problem, it is necessary to keep the color of the signal column in memory in order to be able to select objects of the correct color, even when the column is out of sight. However, in the course of preliminary experiments, it turned out that a reactive agent without memory is able to learn a strategy in which it will always keep the column in sight, even if the room is filled with walls blocking the view from many angles. Therefore, we created a new version of the environment, where the column disappears after the 45th step (see Figure \ref{fig:all_envs}.3.) and the number of walls in the room is significantly reduced. Thus, the subtask of memorization became harder in comparison with the original version, and the subtask of navigation in the environment was simplified. It is also worth noting that the agent does not receive information about the current health or the rewards received, since these observations actually provide the same information as the color of the column.

\section{T-Maze Scaling Experiments.}
\label{sec:long_tmaze}
\begin{figure*}[h!]
\begin{center}
\subfloat[\label{fig:long_tmaze_policy}]{\includegraphics[width=0.49\columnwidth]{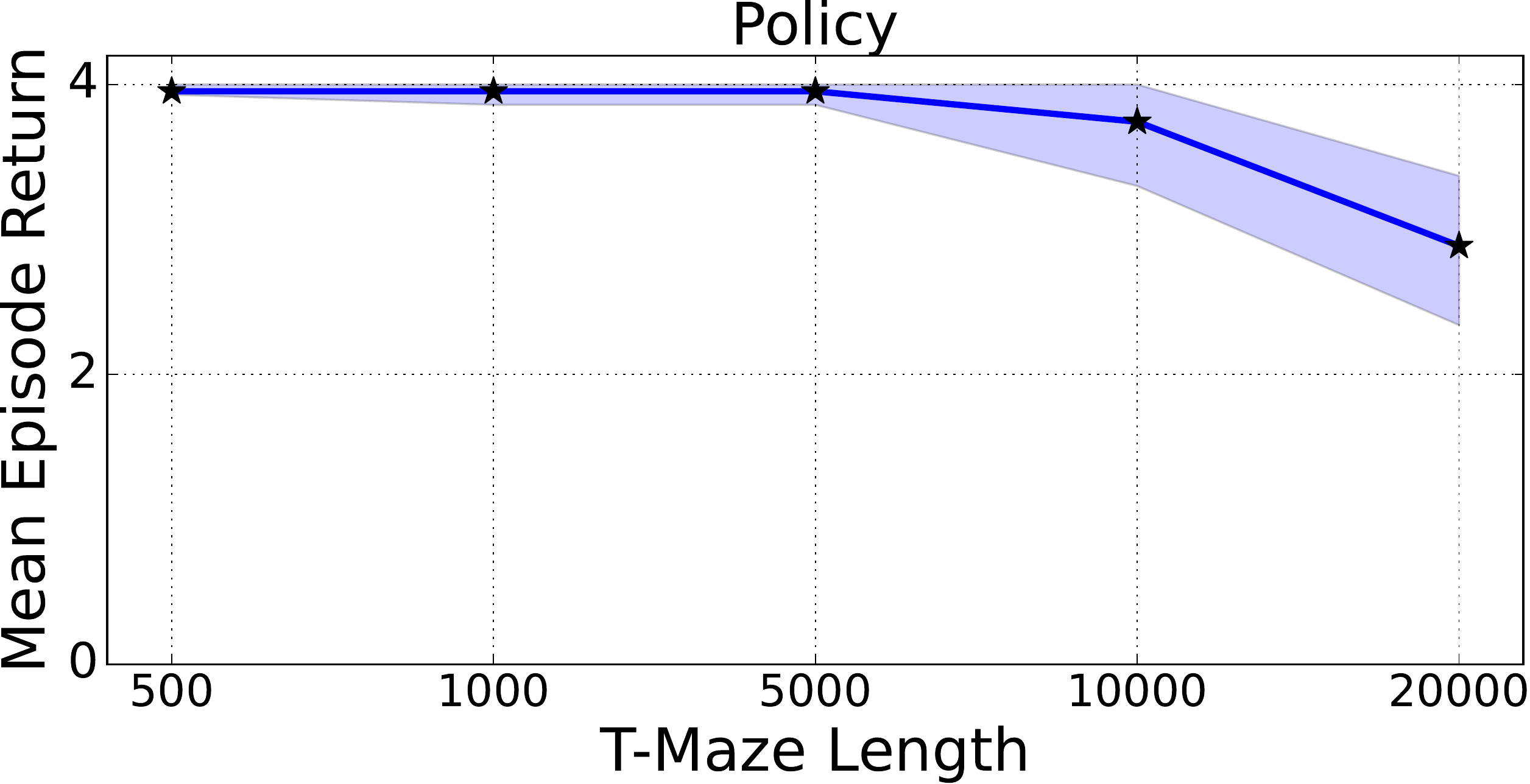}} 
\hfill
\subfloat[\label{fig:long_tmaze_mem}]{\includegraphics[width=0.49\columnwidth]{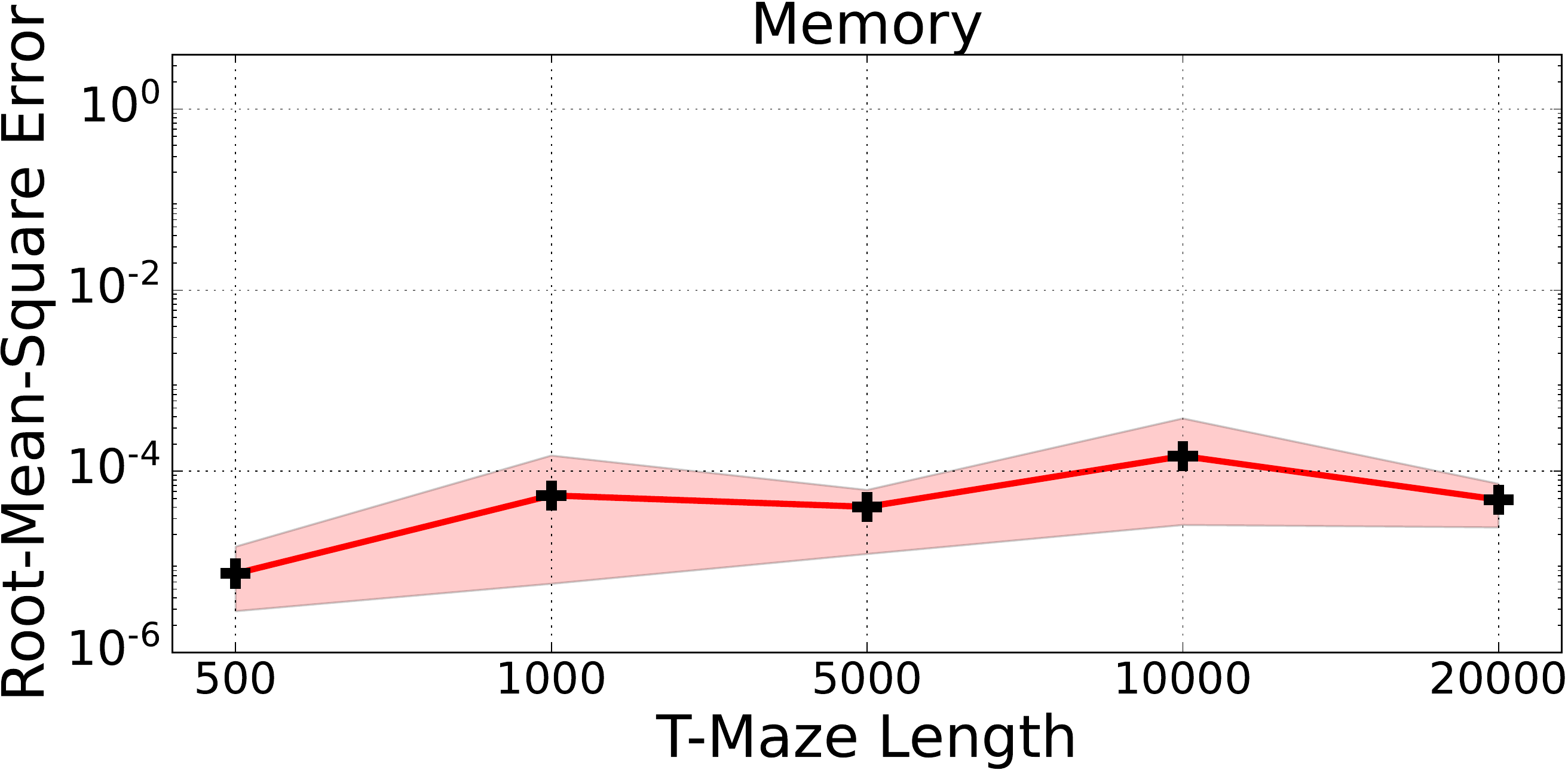}}  

\caption{
\textbf{T-Maze Scaling Experiments.}
\textbf{a)} Final PPO+MemUP performance with respect to corridor length in Noisy T-Maze. Each point is averaged over 3 runs.
\textbf{b)} Final memory and predictor quality with respect to corridor length in Noisy T-Maze. Each point is averaged over 3 runs.  
}



\label{fig:long_tmaze}
\end{center}
\end{figure*}

In this section, we test MemUP's scaling ability in the T-maze-LNR environment. For each version of the environment (with lengths 500, 1000, 5000, 10000, 20000)  we train \texttt{PPO+MemUP} from scratch  in 3 separate runs. Picture \ref{fig:long_tmaze_policy} shows average episodic return after 500 training steps of PPO in the respective environment. Picture \ref{fig:long_tmaze_mem} shows the root mean squared error of the MemUP memory and predictor modules after 200 training epochs. The evaluation of the memory is conducted in the same way as in Section \ref{sec:ablation_study}. The memory module is trained with BPTT rollout $r=20$ in all runs.

\section{Open Source Implementation}\label{sec:code}
Implementation for MemUP can be found at: \url{https://github.com/griver/memup} 

\section{Space Complexity for MemUP}
\label{sec:space_complexity}
Time complexity for MemUP would be the same as for RNNs trained with TBPTT. In our method we focus on space complexity, as MemUP allows RNNs to learn long dependencies with short rollouts.

Space complexity for MemUP is $O(k\frac{r}{f} + r)$, where:
\begin{itemize}
    \item $r$ - rollout length;
    \item $f$ - prediction frequency (you can predict from any intermediate step inside a single rollout);
    \item $k$ - number of prediction targets.
\end{itemize}

In all experiments (excluding Section \ref{sec:ablation_study}) frequency  (long-term predictions are made at the end of each rollout), therefore the space complexity is only $O(k+r)$. We would like to note that we recommend using $f < r$ in longer rollouts as it saves from the vanishing gradient problem inside a single rollout.

\section{Pseudocode for Memory Training}
\label{sec:pseudocode}
This section provides a simplified version of the memory training algorithm with feedforward predictor $q_\phi$.

\begin{algorithm}[h!]
\caption{Recurrent Memory Training}
\label{alg:memup_rnn}
\begin{algorithmic}
  \STATE {\bfseries Input:} inputs $x$, targets $y$, $r$, $K$, $d_\psi$, $q_\phi$, $g_\theta$ 
  \vspace{1mm}
  \STATE Train detector $d_{\psi}$ on the sequence data $x$ and $y$
  \vspace{1mm}
  \STATE Get uncertainty estimates $s$ for each step using $d_\psi$
  \vspace{1mm}
  \STATE $t = 0$ and $m = \texttt{None}$ 
  \vspace{1mm}
  \WHILE{$t \leq T$}
  \vspace{1mm}
    \FOR{$i=t$ {\bfseries to} $t+r$} 
    \STATE $m = g_{\theta}(x_i, m)$
    \ENDFOR
    \vspace{1mm}
    \STATE $t = \min(T, t + r)$
    \vspace{1mm}
    \STATE Using $s[t:T]$ sample a set $\mathcal{U}_t$ of $K$ elements with highest uncertainty  
    \vspace{1mm}
    \STATE Optimize CE loss from eq. 5: $\min_{\theta, \phi} \sum_{k \in \mathcal{U}_t} \left[ - \log q_{\phi}(y_k|m, x_k)\right]$
    \vspace{1mm}
    \STATE \texttt{stop\_gradient}($m$) 
  \vspace{1mm}
  \ENDWHILE
\end{algorithmic}
\end{algorithm}

\end{document}